\documentclass[10pt,twocolumn,letterpaper]{article}

\usepackage{iccv}              

\usepackage{lineno}

\definecolor{iccvblue}{rgb}{0.21,0.49,0.74}
\usepackage[pagebackref,breaklinks,colorlinks,allcolors=iccvblue]{hyperref}
\usepackage{multirow} 
\usepackage{tabularx}

\def\paperID{12573} 
\def\confName{ICCV}
\def\confYear{2025}

\title{Joint Semantic and Rendering Enhancements in 3D Gaussian Modeling with Anisotropic Local Encoding}

\author{
Jingming He\textsuperscript{1} \qquad \quad
Chongyi Li\textsuperscript{2} \qquad \quad
Shiqi Wang\textsuperscript{1} \qquad \quad
Sam Kwong\textsuperscript{3,*} \\
\textsuperscript{1}City University of Hong Kong, Hong Kong SAR, China \\
\textsuperscript{2}Nankai University, Tianjin, China   \quad   \textsuperscript{3}Lingnan University, Hong Kong SAR, China \\
{\tt\small jingmhe3-c@my.cityu.edu.hk \quad lichongyi@nankai.edu.cn \quad shiqwang@cityu.edu.hk \quad samkwong@ln.edu.hk} 
}

\begin{document}
\maketitle

\vspace{-0.07in}

\begingroup
\renewcommand\thefootnote{}\footnotetext{* Corresponding author.}
\endgroup

\begin{abstract}
Recent works propose extending 3DGS with semantic feature vectors for simultaneous semantic segmentation and image rendering.
However, these methods often treat the semantic and rendering branches separately, relying solely on 2D supervision while ignoring the 3D Gaussian geometry.
Moreover, current adaptive strategies adapt the Gaussian set depending solely on rendering gradients, which can be insufficient in subtle or textureless regions.
In this work, we propose a joint enhancement framework for 3D semantic Gaussian modeling that synergizes both semantic and rendering branches.
Firstly, unlike conventional point cloud shape encoding, we introduce an anisotropic 3D Gaussian Chebyshev descriptor using the Laplace–Beltrami operator to capture fine-grained 3D shape details, thereby distinguishing objects with similar appearances and reducing reliance on potentially noisy 2D guidance.
In addition, without relying solely on rendering gradient, we adaptively adjust Gaussian allocation and spherical harmonics (SH) with local semantic and shape signals, enhancing rendering efficiency through selective resource allocation.
Finally, we employ a cross-scene knowledge transfer module to continuously update learned shape patterns, enabling faster convergence and robust representations without relearning shape information from scratch for each new scene.
Experiments on multiple datasets demonstrate improvements in segmentation accuracy and rendering quality while maintaining high rendering frame rates.
\end{abstract}

\section{Introduction}
\label{sec:intro}

3D scene understanding is critical for applications such as robotics navigation and manipulation, where decision-making relies on dense semantic maps from arbitrary viewpoints. Methods based on single-view segmentation or multi-view image fusion often suffer from view inconsistency, while point cloud-based approaches are hindered by inherent sparsity that impedes dense 2D predictions~\cite{guo2024semantic}.

\begin{figure}
  \centering
  \includegraphics[width=0.48\textwidth]{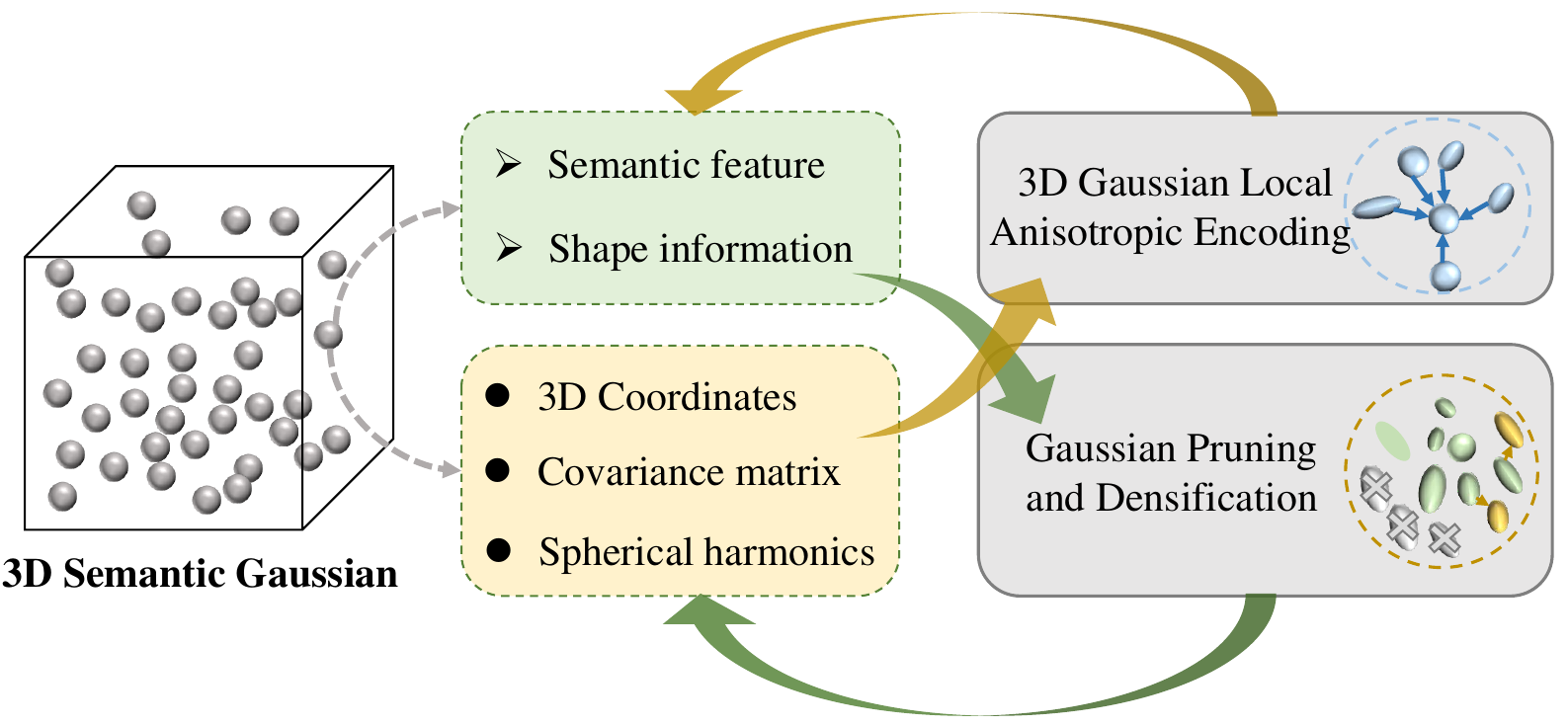}
  \caption{Schematic of our joint enhancement framework in 3D semantic Gaussian. The gold arrow denotes the local anisotropic encoding pathway, leveraging each Gaussian’s intrinsic properties and shape details for semantic refinement. The green arrow represents Gaussian pruning and densification, guided by semantic and shape cues to dynamically add or remove Gaussians and adjust spherical harmonics (SH) levels. This joint semantic-rendering framework enables a more integrated enhancement of semantic understanding and rendering.}
  \label{fig0}
  \vspace{-0.18in}
\end{figure}

Recent works propose extending 3D Gaussian Splatting (3DGS)~\cite{kerbl20233d} with an additional semantic feature vector~\cite{ye2024gaussian,zhou2024feature,zhou2024hugs,choi2024click,shi2024language}, showing the capability to generate dense 2D predictions from any viewpoint. This also allows for simultaneous image rendering, broadening technological applications in fields such as augmented/virtual reality and egocentric perception~\cite{gu2024egolifter}. However, such methods predominantly focus on using pre-trained 2D models to supervise semantic learning, neglecting the influence of the intrinsic structure of the Gaussian representation itself. Moreover, the potential of semantic information to refine processing in the 3D Gaussian rendering branch remains underexplored. Motivated by these observations, we investigates a joint enhancement framework for both semantic and rendering branches in 3D semantic Gaussian modeling, where each component mutually reinforces the other to achieve more precise scene understanding and efficient rendering.

Inspired by semantic-enhanced NeRF~\cite{wang2022dm,zhi2021place,kerr2023lerf,siddiqui2023panoptic,chen2024panoptic}, some methods embed semantic feature vectors into 3D Gaussian representations, rendering semantic features to enable the model to discriminate semantic categories~\cite{peng20243d,liao2024clip,huang2024gaussianformer,qiu2024language,qin2024langsplat,guo2024semantic,li2024versatilegaussian,zhou2024feature}. The rendered feature is supervised through 2D feature distillation methods~\cite{zhou2024feature}, direct pixel-wise 2D supervision~\cite{ye2024gaussian,zhou2024hugs}, or contrastive learning with 2D masks~\cite{choi2024click}. These methods, rely solely on 2D supervision, often missing the rich geometric cues of the 3D domain. This can cause inconsistencies in occluded or ambiguous regions, where unseen scene parts are not accurately captured. Our method, by contrast, encodes local 3D shapes in an anisotropic manner, thus enhancing consistency across views and minimizing dependence on noisy 2D supervision. In terms of rendering enhancements, numerous methods have been proposed to improve quality and efficiency since the introduction of 3DGS. Among these, a branch of approaches learning introduces additional Gaussians in regions exhibiting large residual errors~\cite{wang2024end,rota2024revising,lu2024scaffold},  while others aim to reduce storage overhead by quantizing attributes such as Spherical Harmonics (SH) and covariance or selectively removing redundant Gaussians~\cite{girish2024eagles,cheng2024gaussianpro,lee2024compact}. However, these methods depend solely on local image gradients or fixed thresholds to trigger cloning, splitting, and pruning. Such strategies can be insufficient in regions lacking informative gradients—like textureless surfaces or areas with subtle geometry~\cite{rota2024revising}. We address these challenges by integrating high-level semantic features with our proposed anisotropic shape descriptors, guiding both Gaussian allocation and SH pruning adaptively.

Specifically, our method is designed to jointly enhance both the semantic and rendering branches within 3D semantic Gaussian modeling, as shown in Fig.~\ref{fig0}. In the semantic branch, we focus on enhancing the attached semantic feature vector by incorporating detailed 3D local shape information. To extract directional local shape features, we develop an anisotropic 3D Gaussian Chebyshev descriptor by leveraging Gaussian covariance with anisotropic Laplace–Beltrami operator (ALBO)~\cite{andreux2014anisotropic}. The ALBO has been shown to effectively capture directional variations, such as principal curvature directions, capable of distinguishing adjacent regions that may appear similar but differ in spatial orientations\cite{andreux2014anisotropic,li2020shape,liu2023anisotropic}. We integrate it to capture fine-grained surface details and directional local shape characteristics, essential for differentiating objects with similar color or texture but distinct shapes.
In the rendering branch, we use local semantic and shape cues to assist in guiding the adjustment of the 3D Gaussian set and the selection of spherical harmonics (SH) levels. Regions identified as semantic categories characterized by complex details or exhibiting complex geometry can be allocated a higher density of Gaussians, and vice versa. Moreover, recognizing that many shape patterns and appearance cues are shared across scenes, we have implemented a cross-scene knowledge transfer module based on the core knowledge space (CKS)~\cite{sun2023decoupling}. This module enables faster convergence and more robust representations, allowing the model to avoid relearning shape information from scratch for each new scene.

In summary, our \textbf{contributions} are as follows:

\begin{itemize}
    \item For the semantic branch, we introduce an anisotropic 3D Gaussian local encoding module based on the Laplace–Beltrami operator, enhancing semantic discrimination and reducing reliance on noisy 2D supervision.
    \item For the rendering branch, we adaptively adjust the 3D Gaussian set and spherical harmonics levels using local semantic and shape cues, improving rendering efficiency and visual fidelity.
    \item We perform joint optimization of these branches to leverage their complementary information, and employ a cross-scene knowledge transfer module that strengthens representations by reusing learned shape patterns across multiple scenes.
\end{itemize} 

\section{Related Work}

\subsection{Adaptive Strategies for 3D Gaussian Splatting}
Following the breakthrough of Neural Radiance Fields (NeRF)~\cite{mildenhall2021nerf}, 3DGS~\cite{kerbl20233d} has emerged as a fast, explicit method that leverages anisotropic Gaussians and efficient rasterization. Several works have proposed extensions to improve 3D-GS~\cite{wan2024superpoint,yan2024multi,zhang2024fregs,mihajlovic2024splatfields}. A branch of approaches employs adaptive strategies to adjust Gaussian distribution, selectively adding or removing Gaussians based on optimization signals. For example, some methods learn pruning masks or introduce additional Gaussians in regions exhibiting large residual errors~\cite{wang2024end,rota2024revising,lu2024scaffold}. Scaffold-GS~\cite{lu2024scaffold} employs an anchor-based strategy to dynamically grow Gaussians. Others reduce storage overhead by quantizing attributes such as color and covariance or selectively removing redundant Gaussian~\cite{girish2024eagles,cheng2024gaussianpro,lee2024compact}. Wang~\textit{et al.}~\cite{wang2024end} propose an adaptive mechanism that prunes high-order SH terms for regions with simpler reflectance characteristics. However, existing methods depend solely on local image gradients or fixed thresholds to trigger cloning, splitting, and pruning. In contrast, our method integrates high-level semantic features with anisotropic shape descriptors to guide both Gaussian allocation and SH pruning adaptively.

\subsection{Semantic Integration in 3D Gaussian Splatting}

\begin{figure*}
  \centering
  \includegraphics[width=0.97\textwidth]{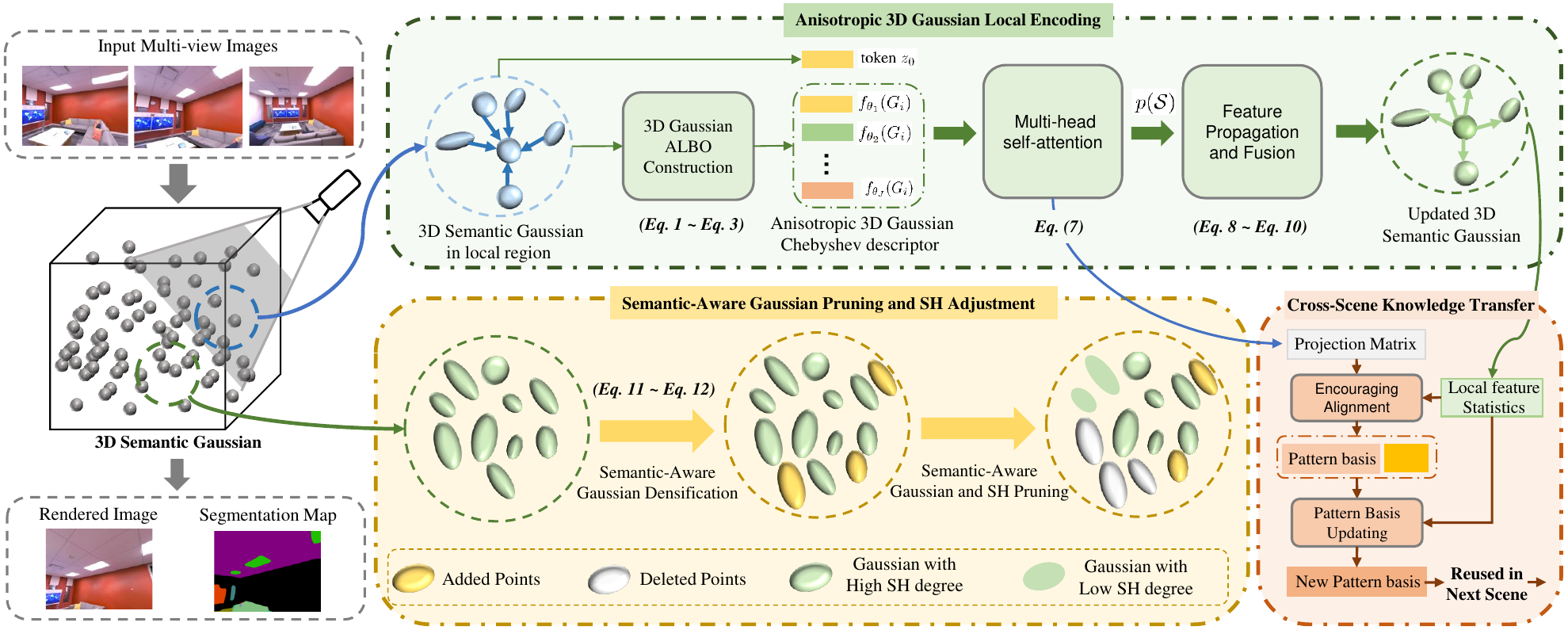}
  \caption{Overview of our proposed pipeline. \textbf{In the anisotropic 3D Gaussian local encoding}, we select multiple local regions within a view frustum. Each local Gaussian is processed with the ALBO to form Chebyshev descriptors, which then enter a transformer-based module to aggregate local geometry information. The refined features are propagated and fused back into individual Gaussians, updating their semantic vectors.  \textbf{In the semantic-aware Gaussian pruning and SH adjustment}, new Gaussians are introduced in areas with high gradients to enhance coverage, while semantic-aware pruning eliminates redundant Gaussians and adjusts SH levels. Finally, we maintain a pattern basis to conduct  \textbf{cross-scene knowledge transfer}. Guided by local feature statistics, scene-specific projection matrices are aligned with this basis. When residuals and shape complexity surpass a threshold, the basis is updated to accommodate new patterns. This basis is stored and reused for future scenes.}
  \label{fig:short}
  \vspace{-0.12in}
\end{figure*}

Previous works have extended NeRF to incorporate semantic information~\cite{wang2022dm,zhi2021place,kerr2023lerf,siddiqui2023panoptic,chen2024panoptic}. Recently, some methods incorporate semantic feature vectors that embed language information into each Gaussian, thereby enabling the model to discriminate between semantic categories~\cite{peng20243d,liao2024clip,huang2024gaussianformer,qiu2024language,qin2024langsplat,guo2024semantic,li2024versatilegaussian,zhou2024feature}. Alternatively, other approaches derive segmentation masks or instance boundaries from 2D cues (e.g., from SAM), primarily partitioning the scene into distinct regions or objects~\cite{gu2024egolifter,choi2024click}. Moreover, another family of methods directly projects 2D semantic predictions (e.g., segmentation masks or CLIP features) onto a pre-constructed 3D Gaussian scene~\cite{shen2024flashsplat,chen2024gaussianeditor}. Nevertheless, despite its computational efficiency, this strategy suffers from view-dependent inconsistencies that often yield conflicting labels in 3D. Additionally, a group of studies concentrates on accelerating the processing of semantic Gaussians~\cite{zuo2024fmgs,ye2024gaussian,shi2024language}. Shi~\textit{et al.}~\cite{shi2024language} introduces a quantization scheme to compress and integrate semantic language features into 3D Gaussians. Existing methods, rely solely on 2D supervision, often overlooking the rich geometric cues inherent in the 3D domain. Unlike these methods, we encode the local 3D shape in an anisotropic manner, directly capturing fine-grained directional features that reflect an object’s intrinsic geometry, which reduces reliance on potentially noisy 2D supervision.


\section{Methodology}

We propose a joint enhancement framework for 3D semantic Gaussian modeling that integrates both semantic and rendering branches, as illustrated in Fig.~\ref{fig:short}. First, we introduce an anisotropic 3D Gaussian local encoding to enrich the semantic feature vectors: each local region within a view frustum is processed with the ALBO to form Chebyshev descriptors, then a transformer-based module aggregates and propagates these features back to individual Gaussians. Next, the semantic-aware Gaussian pruning and SH adjustment module process adaptively duplicates or removes Gaussians according to local gradients and semantic cues, while adjusting SH levels for efficient rendering. Finally, we implement cross-scene knowledge transfer by maintaining a pattern basis that captures recurrent shape patterns, facilitating faster convergence and robust representations in new scenes.

Building on the preliminary concepts of 3DGS, each Gaussian is parameterized by a center $\boldsymbol{\mu} \in \mathbb{R}^3$, an anisotropic covariance matrix $\boldsymbol{\Sigma} \in \mathbb{R}^{3\times3}$, a color vector $\mathbf{c} \in \mathbb{R}^3$, and an opacity value $\alpha\in [0,1]$. For a detailed 3DGS rendering process, please refer to the supplementary material. Building on recent work~\cite{ye2024gaussian,zhou2024feature,zhou2024hugs,choi2024click,shi2024language}, we further attach a semantic feature vector \(\mathbf{f}\in\mathbb{R}^D\) to each Gaussian, forming an extended parameter set $\Theta = \{\boldsymbol{\mu},\, \mathbf{R},\, \mathbf{S},\, \alpha,\, \mathbf{c},\, \mathbf{f}\}.$ The rendered semantic feature map is computed analogously to color blending:$F_s(\mathbf{u}) = \sum_{i=1}^{N} \mathbf{f}_i\,\alpha_i\,T_i$, with $T_i= \prod_{j=1}^{i-1}(1-\alpha_j)$.

\subsection{Anisotropic 3D Gaussian Local Encoding with Laplace Beltrami Operators}

In this section, our objective is to enhance the semantic feature vector $\mathbf{f}$ attached to each Gaussian by incorporating 3D local shape information. We explicitly encode this information by leveraging 3D Gaussian anisotropic covariance with anisotropic Laplace–Beltrami operator (ALBO)~\cite{andreux2014anisotropic}. This enables the model to capture fine-grained surface details, such as curvature and orientation~\cite{andreux2014anisotropic,li2020shape,liu2023anisotropic}.

\subsubsection{Anisotropic 3D Gaussian Chebyshev descriptor}

To reduce computational load and ensure that subsequent local shape encoding is guided by view-specific semantic cues, we restrict the processing spatial domain to the camera's current viewing frustum (See supplementary material for details). We gather the points in this frustum region, then perform iterative farthest point sampling to define local centers $\mathcal{C}$ and their neighborhoods $N(c^{(i)}) = \{ p\in\mathcal{P} \mid \|p-c^{(i)}\|\le r \}$. In each region, we extract proposed Anisotropic 3D Gaussian Chebyshev descriptors. We first construct a graph over the Gaussian centers within the local spatial domain and compute a discrete 3D Gaussian ALBO. The spectral components are then obtained via eigen-decomposition, and Chebyshev polynomials~\cite{liu2023anisotropic} are used to generate spectral descriptors for each Gaussian. Specifically, each 3D Gaussian is represented by its center $c_i\in\mathbb{R}^3$, an anisotropic covariance matrix $\Sigma_i\in\mathbb{R}^{3\times3}$ (assumed positive definite), and an opacity $\alpha_i\in[0,1]$. We first decompose $\Sigma_i$ via eigen-decomposition: $\Sigma_i = R_i\,\Lambda_i\,R_i^T,\quad \Lambda_i=\mathrm{diag}(\lambda_{i1},\,\lambda_{i2},\,\lambda_{i3})$, where $R_i\in SO(3)$ is a rotation matrix and each $\lambda_{ik}>0$ represents the scale along a principal direction. To capture local anisotropic properties, we define a metric matrix
\begin{equation}
M_i = R_i\,\mathrm{diag}\Bigl(\frac{1}{1+\beta\,\lambda_{i1}},\,\frac{1}{1+\beta\,\lambda_{i2}},\,\frac{1}{1+\beta\,\lambda_{i3}}\Bigr)\,R_i^T,
\end{equation}
with $\beta>0$. We then construct a graph $\mathcal{G}=(V,E)$ on the set of Gaussian centers $\{c_i\}$ and define the edge weight for neighboring Gaussians $G_i$ and $G_j$ as
\begin{equation}
w_{ij} = \exp\!\left(-\frac{(c_i-c_j)^T\,\bar{M}_{ij}\,(c_i-c_j)}{\sigma^2}\right),
\end{equation}
where $\bar{M}_{ij} = \frac{1}{2}(M_i+M_j)$ and $\sigma>0$. With normalization $a_i = \sum_{j\in \mathcal{N}(i)} w_{ij}$, the discrete 3D Gaussian ALBO is defined by
\begin{equation}
L_{ij} =
\begin{cases}
-\dfrac{w_{ij}}{a_i}, & i\neq j,\\[1mm]
1, & i=j.
\end{cases}
\end{equation}
We solve the eigenproblem $L\,\phi_k = \lambda_k\,\phi_k$ for $k=1,\dots,K$ and scale the eigenvalues into $[-1,1]$ via $\tilde{\lambda}_k = \tfrac{2\lambda_k}{\lambda_{\max}}-1$,
with $\lambda_{\max}$ being the maximum among the selected $K$. The $d$th-order Chebyshev descriptor for Gaussian $G_i$ is then defined as
\begin{equation}
g_i^d = \sum_{k=1}^{K} T_d\bigl(\tilde{\lambda}_k\bigr) \bigl(\phi_k(i)\bigr)^2,\quad d=0,1,\dots,D,
\end{equation}
where $T_d(\cdot)$ is the Chebyshev polynomial of degree $d$ and $D$ is the maximum order. To incorporate directional sensitivity, the local metric is rotated by a set of angles $\{\theta_j\}_{j=1}^{J}$, yielding rotated metrics $M_i^{\theta_j} = R_{\theta_j}\,M_i\,R_{\theta_j}^T$, from which corresponding Laplacians $L^{\theta_j}$ and eigenpairs $\{\tilde{\lambda}_k^{\theta_j},\phi_k^{\theta_j}\}$ are computed. The Chebyshev descriptor under rotation $\theta_j$ is given by
\begin{equation}
f_{\theta_j}(G_i) = [\, g_{\theta_j}^0(i),\,g_{\theta_j}^1(i),\,\dots,\,g_{\theta_j}^D(i)]^T,
\end{equation}
where $g_{\theta_j}^d(i) = \sum_{k=1}^{K} T_d\bigl(\tilde{\lambda}_k^{\theta_j}\bigr) \bigl(\phi_k^{\theta_j}(i)\bigr)^2$. Finally, concatenating the descriptors across all orientations yields the anisotropic 3D Gaussian Chebyshev descriptor for $G_i$:
\begin{equation}
\mathbf{f}(G_i)= \mathrm{concat}\{ f_{\theta_1}(G_i),\, f_{\theta_2}(G_i),\,\dots,\, f_{\theta_J}(G_i) \}\in\mathbb{R}^{Q},
\end{equation}
where $Q=J\times (D+1)$ denotes the overall descriptor dimension.

\subsubsection{Local Encoding and Feature Propagation}
\label{sec:local_encoding}

In this part, we encode the obtained local anisotropic descriptors derived from 3D Gaussians using a transformer-based module and then propagate these local encodings to individual points via a weighted fusion mechanism.

\textbf{Transformer-based Local Encoding.} Given a local region containing $n$ 3D Gaussians $\mathcal{S} = \{G_i\}_{i=1}^{n}$, each Gaussian is first characterized by its anisotropic Chebyshev descriptor $\mathbf{f}(G_i) \in \mathbb{R}^{Q}$ (as obtained previously) and by its center $c_i\in\mathbb{R}^3$. In order to embed spatial context, we apply a positional encoding function $\mathrm{PE}(\cdot)$ to each center and combine it with the descriptor: $z_i = \mathbf{f}(G_i) + \mathrm{PE}(c_i),\quad i=1,\dots,n$.

A learnable token $z_0\in \mathbb{R}^{Q}$ is prepended to form the input sequence $\mathbf{z} = [z_0; z_1; \dots; z_n] \in \mathbb{R}^{(n+1)\times Q}.$
This sequence is processed by a transformer-based encoder that exploits multi-head self-attention to facilitate feature interactions among local Gaussians. Specifically, the input $\mathbf{z}$ is projected via learnable weight matrices $W_q,W_k,W_v\in\mathbb{R}^{Q\times d}$ into queries, keys, and values: $q = \mathbf{z}\,W_q,\quad k = \mathbf{z}\,W_k,\quad v = \mathbf{z}\,W_v.$
For each of the $m$ attention heads, the head-specific output is computed as $y'_h = \sum_{j=0}^{n} \mathrm{softmax}\Bigl(\frac{q_h \cdot k_j^T}{\sqrt{d}}\Bigr)v_j,\quad h=1,\dots,m$, where $q_h$ denotes the query vector for the $h$th head and $d$ is the projection dimension. The outputs of all heads are then concatenated:
\begin{equation}
y = \mathrm{Concat}(y'_1, y'_2, \dots, y'_m).
\end{equation}
After each self-attention and feed-forward block, the transformer outputs an updated sequence. The learnable token at the output, denoted by $y_0^{(L)}$ after $L$ layers, is passed through a multilayer perceptron (MLP) to obtain a global local encoding $p(\mathcal{S})$, which encapsulates the aggregated anisotropic geometric information in the local region.

\textbf{Local-to-Point Feature Propagation and Fusion.} Individual points within the region may experience varying influences due to their spatial positions and the extent of Gaussian spread. To account for this, we propagate the region encoding to each point using a distance- and covariance-aware weighting scheme. For a local center $c^{(i)}$ with associated encoding $p(\mathcal{S}_i)$ and covariance matrix $\Sigma_i$, any point $q$ in the corresponding neighborhood $N(c^{(i)})$ is assigned a weight
\begin{equation}
\omega_i(q) = \exp\!\left(-\frac{(q-c^{(i)})^T\,\Sigma_i^{-1}\,(q-c^{(i)})}{\tau}\right),
\end{equation}
where $\tau>0$ is a temperature parameter controlling the decay rate. In the case where a point $q$ falls within multiple neighborhoods, we compute its aggregated shape feature by a soft assignment:
\begin{equation}
f^{\mathrm{shape}}(q) = \sum_{i:\,q\in N(c^{(i)})} \frac{\omega_i(q)}{\sum_{j:\,q\in N(c^{(j)})} \omega_j(q)}\, p\bigl(\mathcal{S}_i\bigr).
\end{equation}
Simultaneously, each point $q$ is endowed with a semantic feature $f^{\mathrm{sem}}(q)\in\mathbb{R}^{d_m}$. These two components are concatenated to form a joint feature $\mathbf{u}(q) = [f^{\mathrm{shape}}(q), f^{\mathrm{sem}}(q)]$, and a learnable linear transformation is applied to compute a gating factor $g(q)$. The final point feature is computed by combining the semantic and shape components in a gated manner:
\begin{equation}
f^{\mathrm{final}}(q) = g(q) \odot f^{\mathrm{sem}}(q) + \Bigl(1-g(q)\Bigr) \odot f^{\mathrm{shape}}(q),
\end{equation}
with $\odot$ denoting element-wise multiplication.

\subsection{Semantic-Aware Gaussian Pruning and Spherical Harmonics (SH) Adjustment}

In this part, we use local semantic cues to guide the adjustment of the 3D Gaussian set. Additionally, the selection of SH levels is informed by both these cues and the color complexity derived from local histograms.

\textbf{Gaussian Pruning and Densification.} Each Gaussian is associated with a learnable pruning parameter $\phi_i\in\mathbb{R}$ (consistent with previous methods~\cite{girish2024eagles,rota2024revising,lee2024compact}) and a fused feature $f_i\in\mathbb{R}^d$ (obtained from the previous semantic and shape fusion module). To decide whether a Gaussian should be retained, we first compute a gated version of its feature using a linear layer with sigmoid activation to obtain an updated feature $f'_i$. Next, we form a concatenated vector $\mathbf{u}_i = [f'_i,\, \phi_i] \in \mathbb{R}^{d+1}$ and pass it through a second gating network to yield a hidden representation, which is then linearly mapped to produce a soft mask $\hat{m}_i\in (0,1)$. Thresholding $\hat{m}_i$ using a fixed value $\tau$ with a straight-through estimator (STE) produces the binary mask $m_i\in\{0,1\}$, which is applied to update the Gaussian's scale and opacity as $\tilde{s}_i = m_i \cdot s_i$ and $\tilde{\alpha}_i = m_i \cdot \alpha_i$. A pruning loss, $L_{\mathrm{mask}} = \frac{1}{N}\sum_{i=1}^{N} \hat{m}_i$, encourages suppression of unnecessary Gaussians. Moreover, in regions (computed over a local neighborhood centered at $c^{(i)}$) where the average gradient magnitude
\begin{equation}
\bar{\nabla}(c^{(i)}) = \frac{1}{|N(c^{(i)})|}\sum_{q\in N(c^{(i)})} \|\nabla f(q)\|
\end{equation}
 exceeds a predefined threshold, additional Gaussians are introduced. For a local region with center $c$, we compute the candidate count as
\begin{equation}
N_p = \min\Bigl(N_{\max},\, \alpha_d\cdot \frac{V_{\mathrm{pc}}}{N_{\mathrm{ori}}}\Bigr),
\end{equation}
where $V_{\mathrm{pc}}$ is the point cloud volume, $N_{\mathrm{ori}}$ is the number of original points, $\alpha_d>0$ is a density parameter, and $N_{\max}$ is an upper bound. Candidate points are uniformly sampled within a sphere of radius $r$ centered at $c$; for each candidate $p_{\mathrm{new}}$, the nearest neighbor distance $d(p_{\mathrm{new}},p_{\mathrm{NN}}) = \|p_{\mathrm{new}} - p_{\mathrm{NN}}\|$ is computed, and candidates outside the interval $[d_{\min}, d_{\max}]$ are rejected. Accepted candidates inherit Gaussian attributes from their nearest neighbor and are added to the set, thereby increasing local density where needed.

\textbf{Spherical Harmonics Pruning.} For each Gaussian $i$ and SH degree $l\in\{0,1,2,3\}$, let the SH coefficients be denoted by $c_i^{(l)}\in\mathbb{R}^{(2l+1)\times 3}$ and assign a learnable raw mask parameter $\psi_i^{(l)}\in\mathbb{R}$ (consistent with~\cite{wang2024end}). Let $f_i\in\mathbb{R}^d$ be the fused semantic-shape feature for Gaussian $i$. To quantify local color complexity at degree $l$, we extract the set of color values $\mathcal{X}^{(l)}(\mathcal{R})$ from the local region $\mathcal{R}$ and construct a differentiable soft histogram. Suppose the color range is $[m, M]$, partitioned into $B$ bins of width $\Delta = \frac{M-m}{B}$ with bin centers $c_b = m + \Delta(b+\frac{1}{2})$ for $b=0,\dots,B-1$. The soft histogram for bin $b$ is computed as:
\begin{align}
H^{(l)}(\mathcal{R})[b] &= \sum_{x\in \mathcal{X}^{(l)}(\mathcal{R})}
\Bigl[\varsigma\bigl(\gamma(x-(c_b-\Delta/2))\bigr) \notag \\
&\quad -\varsigma\bigl(\gamma(x-(c_b+\Delta/2))\bigr)\Bigr].
\end{align}
where $\varsigma(z)=\frac{1}{1+e^{-z}}$ denotes the logistic sigmoid function, and $\gamma>0$ is the softness parameter that controls the sharpness of the bin boundaries. The resulting histogram is then embedded via an MLP: $\eta^{(l)}(\mathcal{R}) = \mathrm{MLP}_{H}^{(l)}\Bigl(H^{(l)}(\mathcal{R})\Bigr) \in \mathbb{R}^{d_h}$. For each Gaussian $i$ and SH degree $l$, we form a joint feature vector $\mathbf{v}_i^{(l)} = [f_i,\, \psi_i^{(l)},\, \eta^{(l)}(\mathcal{R})] \in \mathbb{R}^{d+1+d_h}$ and then apply a linear mapping with sigmoid activation to yield a soft mask value $\delta_i^{(l)}\in (0,1)$. Binarizing $\delta_i^{(l)}$ using a threshold $\tau_l$ (via STE) produces $\hat{m}_i^{(l)}\in\{0,1\}$, which is applied to the SH coefficients as $\tilde{c}_i^{(l)} = \hat{m}_i^{(l)}\cdot c_i^{(l)}$. To promote sparsity, we define the SH pruning loss as $L_{\mathrm{SH}} = \frac{1}{N}\sum_{i=1}^{N}\sum_{l=0}^{3} w_l\, \delta_i^{(l)}$, with $w_l=2l+1$. The entire training objective integrates losses from the semantic branch, reconstruction, and semantic-aware Gaussian pruning.

\subsection{Continuous Learning Module for Cross-Scene Knowledge Transfer}

Recognizing that many shape patterns and appearance cues are shared across scenes, we implemented a cross-scene knowledge transfer module to facilitate continuous learning for the transformer module. Based on the core knowledge space (CKS)~\cite{sun2023decoupling}, we dynamically adjust the regularization by calculating local feature statistics and updating the core knowledge space by considering both projection residuals and local shape complexities. Let $W \in \mathbb{R}^{q \times d}$ be a key projection matrix (e.g., one of the transformer’s projection matrices in Sec.~\ref{sec:local_encoding}) for the current scene, and let $B \in \mathbb{R}^{q \times r}$ be the associated pattern basis for the core knowledge space (CKS), with $r < q$ and $B^\top B = I_r$. We project $W$ onto the CKS via $\widehat{W} = B\,B^\top W,$ and define the residual $R = W - \widehat{W}.$ The relative projection error is quantified by the ratio $\rho = \frac{\|R\|_F}{\|W\|_F},$ where a larger $\rho$ indicates that $B$ does not sufficiently capture the information in $W$.

\textbf{Spatially Weighted Regularization.} To encourage $W$ to reside within the CKS, we modulate the regularization strength based on the local feature statistics. Let $\bar{z} = \frac{1}{n+1}\sum_{i=0}^{n} z_i$, where $z_i$ is the local shape feature of a Gaussian (from Sec.~\ref{sec:local_encoding}) and $n$ is the number of Gaussians in the local region. A modulation factor $\alpha$ is computed via a single-layer network taking $\bar{z}$ as input, and the regularization loss is then defined as
\begin{equation}
\mathcal{L}_{\mathrm{reg}}(W) = \alpha \cdot \frac{\|W - B\,B^\top W\|_F^2}{\|W\|_F^2}.
\end{equation}
This term encourages $W$ to align with the CKS, with stronger regularization when the modulated value $\alpha$ is high. We incorporate this loss term into the overall training objective when this module is activated.

\textbf{Pattern Basis Update Strategy.} To decide whether the current basis $B$ should be updated, we combine the residual ratio $\rho$ with a measure of local shape complexity. Each Gaussian $G_i$ is characterized by its anisotropic Chebyshev descriptor $f(G_i)$. For a local region containing $N$ Gaussians, we compute the average descriptor $\bar{f} = \frac{1}{N}\sum_{i=1}^{N} f(G_i)$ and the standard deviation
\begin{equation}
\sigma_f = \sqrt{\frac{1}{N}\sum_{i=1}^{N} \|f(G_i) - \bar{f}\|_2^2}.
\end{equation}
We then define a combined indicator $\rho' = \rho \cdot \bigl(1 + \kappa\, \sigma_f\bigr)$, with $\kappa>0$ as a balancing parameter. When $\rho'$ exceeds a preset threshold $\epsilon>0$, the current pattern basis is deemed inadequate. To update $B$ when $\rho$ is high, we perform SVD on $R$: $R = U\,\Sigma\,V^\top,$ and select the top $r'$ singular vectors such that$\frac{\sum_{i=1}^{r'} \sigma_i^2}{\sum_{i=1}^{\min(q,d)} \sigma_i^2} \ge \eta,$
with $\eta\in (0,1)$ representing the energy retention ratio. The updated basis is then obtained by orthogonalizing the union of $B$ and the selected vectors $U^{(r')}$: $B \leftarrow \operatorname{orth}\Bigl([B,\; U^{(r')}]\Bigr).$ At the end of training, we perform the basis update for each key projection matrix as described above. During inference or new training, the stored pattern basis is reloaded to facilitate shape encoding in the new scene.

\section{Experiments}

\subsection{Datasets and Experimental Setups}

\begin{figure*}
  \centering
  \includegraphics[width=\textwidth]{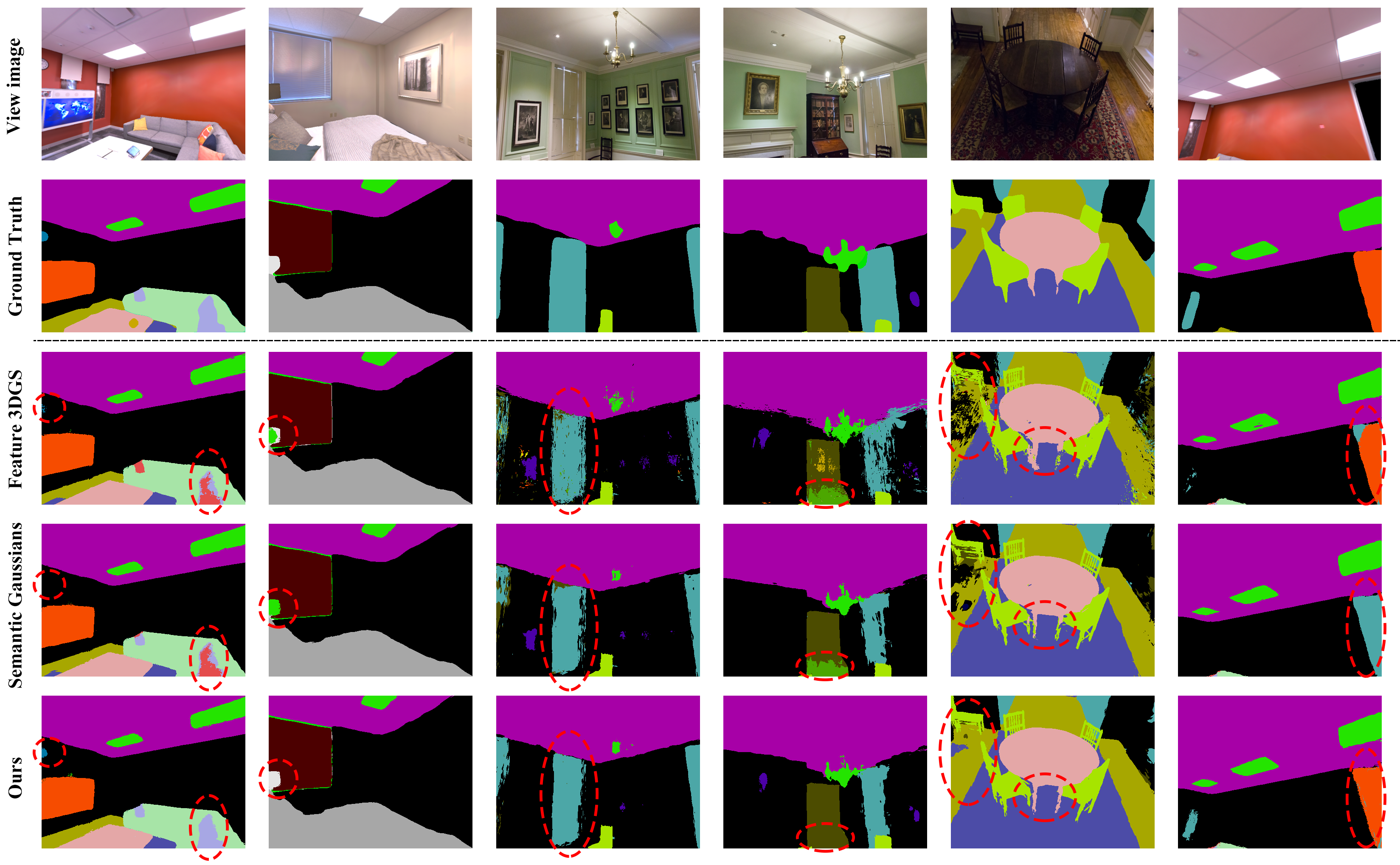}
  \caption{Visualization of semantic segmentation results. From top to bottom displays the View image, Ground Truth, results from Feature 3DGS~\cite{zhou2024feature}, Semantic Gaussians~\cite{guo2024semantic}, and our results.}
  \label{fig:vis1}
  \vspace{-0.08in}
\end{figure*}

\textbf{The Replica Dataset}~\cite{straub2019replica} comprises 18 high-quality, photo-realistic reconstructions of indoor scenes at room and building scales. For supervision, same with~\cite{zhou2024feature}, we apply the pre-trained LSeg model~\cite{li2022language} to generate semantic features from 2D images, and then align the rendering 3D semantic features closely with the corresponding 2D teacher features using L1 loss. \textbf{The Deep Blending}~\cite{hedman2018deep} has a diverse collection of 2630 high-resolution images across 19 real-world scenes. We conduct experiments on the "drjohnson" and "playroom" scenes, consistent with the settings in \cite{kerbl20233d,girish2024eagles,lee2024compact}. The supervision approach is the same as that used in the Replica dataset.  \textbf{The ScanNet dataset}~\cite{dai2017scannet} is a large-scale RGB-D video dataset aimed at 3D scene understanding. We selected 12 scenes from the validation set, in line with \cite{siddiqui2023panoptic,guo2024semantic}. We conducted experiments on both open-vocabulary and closed-set segmentation. The supervision for open-vocabulary segmentation follows the same method as used with the Replica dataset. For closed-set segmentation, same as \cite{siddiqui2023panoptic}, we used pseudo labels for supervision. More details, including optimization and hyperparameters, are provided in the supplementary material.

\begin{table}
  \centering
  \resizebox{0.35\textwidth}{!}{
  \begin{tabular}{l|cc|c}
    \toprule
    {Method} & mIoU  & OA & PSNR \\
    \midrule
    NeRF-DFF~\cite{kobayashi2022decomposing} & 63.6  & 86.4  & 32.85  \\
    Panoptic Lifting~\cite{siddiqui2023panoptic} & 67.2  & -     & - \\
    Feature 3DGS~\cite{zhou2024feature} & 78.2  & 94.3  & 36.20  \\
    \midrule
    Ours  & \textbf{81.1} & \textbf{95.9} & \textbf{36.58} \\
    \bottomrule
  \end{tabular}}
  \vspace{-0.07in}
  \caption{Results of Segmentation and Rendering on the Replica Dataset. Bold values represent the best results.}
  \label{tab:replica}
  \vspace{-0.06in}
\end{table}

\begin{table}
  \centering
   \resizebox{0.48\textwidth}{!}{
  \begin{tabular}{l|c|cc|c}
   \toprule
    {Method} & Method Type & mIoU  & mAcc  & PSNR \\
    \midrule
    OpenSeg~\cite{ghiasi2022scaling} & \multirow{2}[2]{*}{Image-based} & 53.4  & 75.1  & - \\
    LSeg~\cite{li2022language} &       & 56.1  & 74.5  & - \\
    \midrule
    LERF~\cite{kerr2023lerf} & \multirow{2}[2]{*}{NeRF-based} & 31.2  & 61.7  & - \\
    PVLFF~\cite{chen2024panoptic} &       & 52.9  & 67.0  & - \\
    \midrule
    LangSplat~\cite{qin2024langsplat} & \multirow{3}[2]{*}{3DGS-based} & 24.7  & 42.0  & - \\
    Feature3DGS~\cite{zhou2024feature} &       & 59.2  & 75.1  & 26.98  \\
    Semantic Gaussians~\cite{guo2024semantic} &       & 62.0  & 77.0  & 26.93  \\
    \midrule
    Ours  & \multirow{2}[2]{*}{3DGS-based} & 64.3  & 79.7  & 27.35  \\
    Ours (w/ CSKT) &       & \textbf{65.1} & \textbf{80.5} & \textbf{27.37} \\
    \bottomrule
  \end{tabular}}
  \vspace{-0.05in}
  \caption{Results of Open-Vocabulary Segmentation and Rendering on the ScanNet Dataset. Bold values represent the best results.}
  \label{tab:scannetopen}
  \vspace{-0.15in}
\end{table}

\subsection{Main Results}

\begin{figure}
\vspace{-0.08in}
  \centering
  \includegraphics[width=0.48\textwidth]{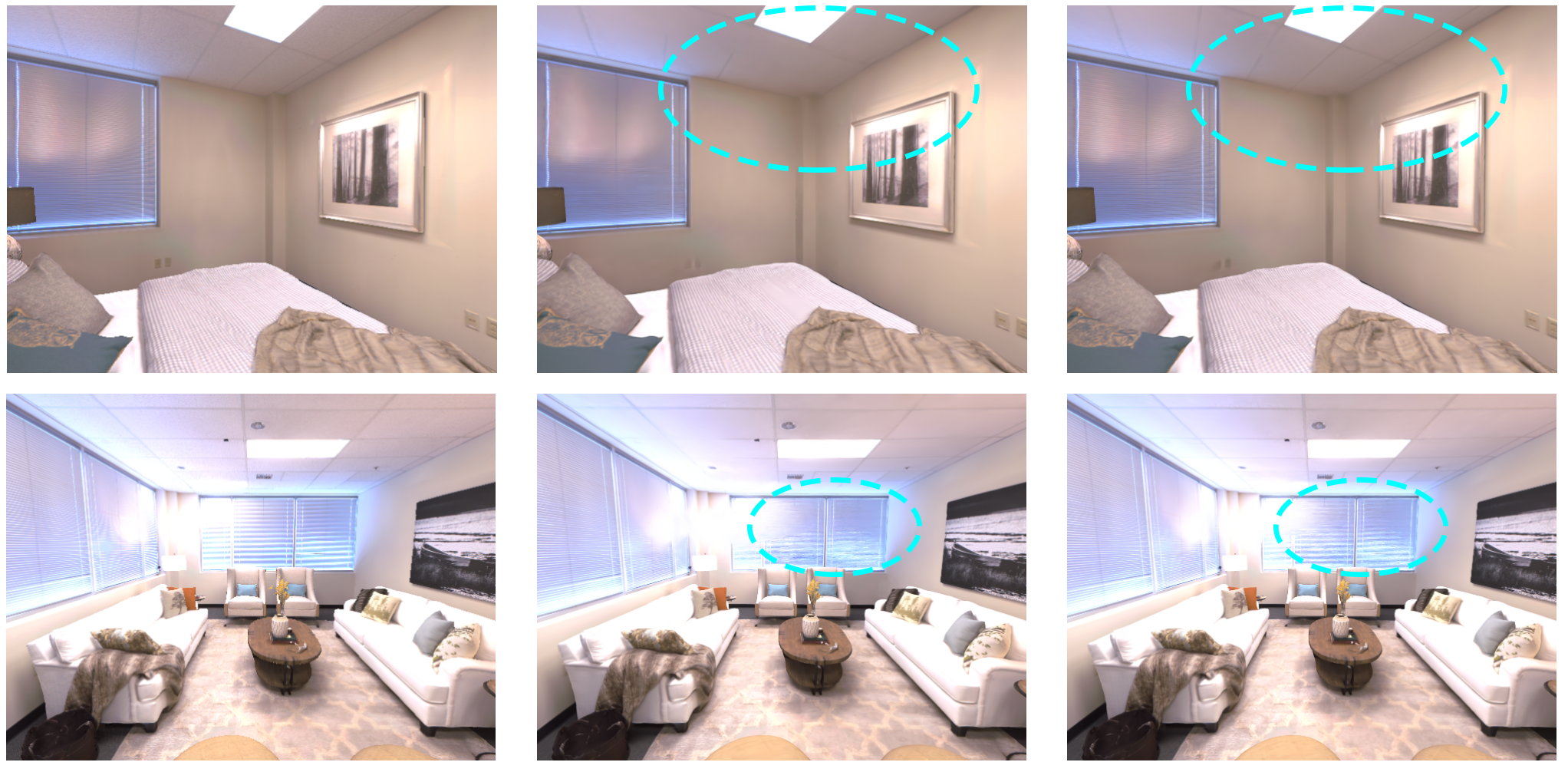}
  \caption{Visualization of rendering results. From left to right shows the Original Scene Image,  the results from Feature 3DGS~\cite{zhou2024feature}, and our results (Under the same number of iterations).}
  \label{fig:vis2}
  \vspace{-0.12in}
\end{figure}

Tables \ref{tab:replica} and \ref{tab:scannetopen} present our results for open-vocabulary segmentation and rendering on the Replica and ScanNet datasets, evaluating using metrics such as \textbf{mIoU}, Overall Accuracy (\textbf{OA}), and \textbf{mAcc}. mIoU measures the average intersection over union, OA evaluates overall pixel accuracy, and mAcc assesses mean class accuracy. Our method achieved the highest segmentation scores with mIoU values reaching 81.1 on Replica and 65.1 on ScanNet. Fig.~\ref{fig:vis1} illustrates the comparative visualization of our segmentation results. Our method demonstrates improved performance in identifying small objects (e.g., lamps and clocks), and distinguishing between objects with similar textures but different shapes, as illustrated by the table legs and chairs in the fifth column. Furthermore, our segmentation results are smoother with fewer noise artifacts. Additionally, the Peak Signal-to-Noise Ratio (\textbf{PSNR}) results also highlight superior rendering quality compared to existing methods. It demonstrates our model's capability to enhance both semantic understanding and rendering performance. Fig.~\ref{fig:vis2} showcases our rendering results, showing enhanced visual quality in areas with complex geometric structures (corners), and in regions with rich details like the textures on blinds and ceilings. This confirms the effectiveness of our adaptive Gaussian adjustment in rendering.

\begin{table}
\vspace{-0.08in}
  \centering
  \resizebox{0.40\textwidth}{!}{
  \begin{tabular}{l|c|cc}
    \toprule
    {Method} & Method Type & mIoU  & PSNR \\
    \midrule
    Mask2Former~\cite{cheng2022masked} & Image-based & 46.7  & - \\
    \midrule
    DM-NeRF~\cite{wang2022dm} & \multirow{3}[2]{*}{NeRF-based} & 49.5  & 28.21  \\
    SemanticNeRF~\cite{zhi2021place} &       & 59.2  & - \\
    Panoptic Lifting~\cite{siddiqui2023panoptic} &       & 65.2  & \textbf{28.50} \\
    \midrule
    Ours  & \multirow{2}[2]{*}{3DGS-based} & 67.4  & 27.49  \\
    Ours (w/ CSKT) &       & \textbf{68.0} & 27.49  \\
    \bottomrule
  \end{tabular}}
  \vspace{-0.05in}
  \caption{Results of Closed-Set Segmentation and Rendering on the ScanNet Dataset. Bold values represent the best results.}
  \label{tab:scannetclose}
  \vspace{-0.17in}
\end{table}

\begin{table}
  \centering
  \setlength{\tabcolsep}{3pt}
  \resizebox{0.48\textwidth}{!}{
  \begin{tabular}{l|cc|ccc|c}
    \toprule
    Method & mIoU$\uparrow$ & OA$\uparrow$ & PSNR$\uparrow$ & SSIM$\uparrow$ & LPIPS$\downarrow$ & FPS$\uparrow$ \\
    \midrule
    3DGS~\cite{kerbl20233d} & -     & -     & 29.55  & 0.900  & \textbf{0.250 } & 109  \\
    EAGLES~\cite{girish2024eagles} & -     & -     & \textbf{29.86} & \textbf{0.910} & \textbf{0.250} & 119  \\
    Compact 3DGS~\cite{lee2024compact} & -     & -     & 29.79  & 0.901  & 0.258  & 152  \\
    Feature 3DGS~\cite{zhou2024feature} & 46.2  & 79.9  & 29.58  & 0.902  & 0.252  & 110  \\
    \midrule
    Ours  & \textbf{50.6} & \textbf{83.1} & \textbf{29.86} & 0.906  & \textbf{0.250} & \textbf{164} \\
    \bottomrule
  \end{tabular}
  }
  \vspace{-0.05in}
  \caption{Results of Segmentation (mIoU, OA), Rendering (PSNR, SSIM, LPIPS), and Speed (FPS) on the Deep Blending Dataset. Bold values represent the best results.}
  \label{tab:db}
  \vspace{-0.18in}
\end{table}

We explore closed-set segmentation and rendering results on the ScanNet dataset.  In addition, we enable our cross-scene knowledge transfer (CSKT) module during ScanNet training, denoted as \textbf{Ours (w/ CSKT)}, since different indoor rooms in ScanNet often share underlying geometric patterns (e.g., similar wall, floor, or furniture shapes). By storing and updating these patterns across different rooms, CSKT allows our model to better adapt when encountering similar geometry or semantic categories in new scenes. Due to the absence of 3DGS-based methods in closed-set tasks on the ScanNet Dataset, we only compared our method against image-based and NeRF-based methods. As shown in Table \ref{tab:scannetclose}, we outperform all image-based and NeRF-based methods in semantic segmentation. However, our rendering quality is constrained by the fundamental differences between 3DGS and NeRF techniques on ScanNet, resulting in lower performance compared to established NeRF-based methods~\cite{wang2022dm,zhi2021place,siddiqui2023panoptic}.

In Table \ref{tab:db}, we test our method on the Deep Blending dataset, a benchmark commonly used for evaluating rendering performance. Our comparisons include methods specifically designed for adaptive Gaussian set management, such as EAGLES~\cite{girish2024eagles} and Compact 3DGS~\cite{lee2024compact}. The reported FPS values were measured on an NVIDIA RTX 3090 GPU. To ensure a fair comparison when measuring frames per second (FPS), particularly because EAGLES and Compact 3DGS focus solely on rendering, we disabled the semantic branch in our tests of Feature 3DGS and our method. We efficiently allocate resources by adaptively adding or removing Gaussians guided by semantic cues and adjusting SH levels, thus avoiding redundancy in less critical areas. Consequently, our method maintains fast rendering speed while achieving competitive results in both rendering and segmentation.

\subsection{Ablation Studies}

\begin{table}
  \centering
  \resizebox{0.42\textwidth}{!}{
  \begin{tabular}{l|cc|cc}
    \toprule
    Method & mIoU  & OA    & PSNR  & FPS \\
    \midrule
    Ours (LEM w/o AGCD) & 48.3  & 80.9  & 29.63  & 154  \\
    Ours (LEM w/o TLE) & 49.2  & 81.7  & 29.78  & 161  \\
    Ours (AGP w/o SG) & 50.4  & 82.8  & \textcolor[rgb]{ 0,  0,  1}{29.86} & 126  \\
    Ours (AGP w/o AD) & 49.8  & 82.5  & 29.50  & \textcolor[rgb]{ 1,  0,  0}{173} \\
    Ours (ASHP w/o SG) & \textcolor[rgb]{ 1,  0,  0}{50.6} & \textcolor[rgb]{ 0,  0,  1}{83.0} & \textcolor[rgb]{ 1,  0,  0}{29.88} & 135  \\
    \midrule
    Ours  & \textcolor[rgb]{ 1,  0,  0}{50.6} & \textcolor[rgb]{ 1,  0,  0}{83.1} & \textcolor[rgb]{ 0,  0,  1}{29.86} & \textcolor[rgb]{ 0,  0,  1}{164} \\
    \bottomrule
  \end{tabular}
  }
  \vspace{-0.07in}
  \caption{Ablation Study Results on the Deep Blending Dataset. Text in red represents the highest values, and blue indicates the second best.}
  \label{tab:as}
  \vspace{-0.13in}
\end{table}

In this section, we conduct several ablation variants of our method on the Deep Blending dataset. Quantitative results are summarized in Table \ref{tab:as}. Firstly, we remove the anisotropic 3D Gaussian Chebyshev descriptor (AGCD) in the local encoding module (LEM), and rely on a simple MLP to encode all points in the local region, denoted as \textbf{Ours (LEM w/o AGCD)}. According to the results, the mIoU drops from 50.6 to 48.3, and OA decreases from 83.1 to 80.9, indicating that AGCD effectively refines semantic features. Additionally, the PSNR also decreases from 29.86 to 29.63, showing that robust shape-informed semantics are crucial to guiding the rendering branch effectively. Then, we replace the transformer-based local encoding (TLE) with MLP, denoted as \textbf{Ours (LEM w/o TLE)}. The performance decreases (e.g., mIoU from 50.6 to 49.2), implying that the transformer-based encoding better captures interactions among local Gaussians.

In the adaptive Gaussian pruning (AGP), we remove semantic guidance (SG) and only use the rendering gradient, denoted as \textbf{Ours (AGP w/o SG)}. We observe that despite no increase in PSNR (29.86), there is a noticeable reduction in efficiency (FPS drops to 126), implying that semantic cues help prune Gaussians more judiciously. Then, we disable adaptive densification (AD) within AGP, so Gaussians cannot be dynamically added, denoted as \textbf{Ours (AGP w/o AD)}. While the rendering speed sees an improvement (173 FPS vs. 164), there is a significant drop in PSNR (reducing to 29.50), and the mIoU also slightly decreases from 50.6 to 49.8. This trade-off highlights that densification is key to handling fine geometry or intricate textures, even though it slightly increases computational overhead. Finally, for Adaptive SH Pruning (ASHP), we remove semantic guidance, denoted as \textbf{Ours (ASHP w/o SG)}. Despite similar PSNR and Segmentation Results, the FPS degrades from 164 to 135. This suggests that without semantic cues to guide SH order, the pruning becomes less targeted.

\section{Conclusion}

In this paper, we propose a joint enhancement framework for 3D semantic Gaussian modeling that bridges both semantic and rendering branches. We employ an anisotropic 3D Gaussian local encoding module to enhance semantic discrimination and reduce dependency on noisy 2D data. By adaptively adjusting Gaussian distributions and SH levels based on semantic and shape cues, our method enhances visual fidelity while keeping efficiency. A cross-scene knowledge transfer module is implemented to update shape patterns continually, leading to more robust representations. Extensive experiments on several datasets demonstrate improvements in segmentation accuracy and rendering fidelity while maintaining high rendering frame rates. 

\section*{Acknowledgment}

This work is partially supported by the Research Grant Council (RGC) of Hong Kong General Research Fund (GRF) under Grant 11200323, the NSFC/RGC JRS Project N\_CityU198/24, the Natural Science Foundation of Tianjin, China (24JCJQJC00020), and the Fundamental Research Funds for the Central Universities (Nankai University, 070-63243143).

{
    \small
    \bibliographystyle{ieeenat_fullname}
    \bibliography{main}
}

\end{document}